\documentclass[sigconf,nonacm]{acmart}
\usepackage{amsmath}
\usepackage{float}
\usepackage{balance}
\usepackage{xurl}
\usepackage{needspace}
\newcommand{\phiadj}{\phi_{\mathrm{adj}}}
\newcommand{\phimax}{\phi_{\mathrm{max}}}
\newcommand{\Sstar}{S^{\star}}
\newcommand{\okA}{\mathrm{ok}_A}
\newcommand{\okB}{\mathrm{ok}_B}

\begin{document}

\title{Quantifying Diversity of Thought: A Predictive Law of Weighted LLM
Ensemble Lift}
\titlenote{This research was funded by NCSC, part of GCHQ. Model inference
was performed on the Isambard-AI supercomputer, provided by UKRI.}

\author{Junade Ali}
\email{jali@turing.ac.uk}
\affiliation{%
  \department{Defence and National Security Programme}
  \institution{The Alan Turing Institute}
  \city{London}
  \country{United Kingdom}
}

\begin{abstract}
This paper provides an experimentally verified formal law for calculating the
uplift that diversity of thought provides in Large Language Model (LLM)
ensembles. From
first principles, we derive an exact decomposition of LLM ensemble lift into
rescue and damage masses, which yields a compact heuristic for calculating
uplift. From this we extract the metrics which predict ensemble performance:
an accuracy-adjusted correctness correlation, $\phiadj$, together with the
accuracy gap and collective accuracy of the pair. We test the law on 767{,}520
inferences from ten open-weight models over two graduate-level science
benchmarks, together with a novel agentic cybersecurity benchmark in which
each model conducts digital-forensics investigations by multi-turn tool use
in a network-isolated sandbox (23{,}520 graded trials including
abstentions); all votes are released openly.
Calibrated once on SuperGPQA at a 40:60 vote split, the heuristic predicts
lift on the calibration set with Spearman's $\rho=0.84$ and, with its
coefficients frozen, transfers to two datasets never used in calibration ($\rho=0.51$ on GPQA Diamond and
$0.84$ on the forensic tasks), whilst the measured swap mass tracks realised
lift with $R^2\ge 0.96$ throughout. Raw $\phi$
has almost no predictive power ($R^2\le 0.09$ throughout); the
accuracy-adjusted $\phiadj$ is markedly superior ($R^2=0.67$ on SuperGPQA), and the heuristic combining
these metrics is the most stable pre-pooling predictor across the three
datasets.
\end{abstract}


\maketitle

\section{Introduction}\label{sec:intro}
Diversity of thought is the oldest argument for combining predictors: a panel
of imperfect voters can outperform its best member whenever the voters err in
different places. For large language models the argument has been borne out
empirically many times over --- sampling a single model repeatedly and taking
a plurality vote reliably raises accuracy~\cite{wang2022selfconsistency}, and
accuracy continues to climb as more sampled agents are
added~\cite{li2024more}. A fast-growing literature now combines \emph{distinct}
models by learned fusion~\cite{jiang2023llmblender}, layered
mixtures~\cite{wang2024mixtureofagents}, cost-driven
cascades~\cite{chen2023frugalgpt} and reward-guided
routing~\cite{lu2024routing}. The common finding is that combination helps ---
benchmark by benchmark, and usually for reasons diagnosed after the fact.

What this literature does not supply is a means of predicting, before any
pooled inference is run, whether adding a particular weaker model to a stronger
one will help or harm, and at what vote weight. The classical study of
ensemble diversity has repeatedly found that structural indicators of
complementarity correlate weakly and unstably with realised
gain~\cite{kuncheva2003measures,shipp2002relationships}, and the recent
human--AI evidence is a caution rather than an encouragement: a meta-analysis
of 370 effect sizes found combinations that on average \emph{underperform}
their best member, with outcomes moderated by task type and the partners'
relative solo strength~\cite{vaccaro2024when}. It remains unknown whether a compact,
interpretable law can separate the helpful pairings from the harmful ones
across tasks as heterogeneous as multiple-choice science and agentic
tool use.

In this paper, we provide such a law, derived from first principles and
verified experimentally. Our contributions are threefold. Firstly, we derive
the \emph{weighted swap law}: an exact, assumption-free decomposition of
weighted two-model ensemble lift into rescue and damage masses, from which we
extract a compact heuristic for calculating uplift and the metrics that
predict ensemble performance --- an accuracy-adjusted correctness
correlation $\phiadj$ (the classical $\phi/\phimax$), alongside the
accuracy gap and collective accuracy of the pair. Secondly, we
construct the evaluation data the question requires, including a novel agentic
cybersecurity benchmark: ten open-weight models each conduct digital-forensics
investigations by multi-turn tool use inside a network-isolated sandbox,
answering 98 questions graded offline against gold answers. Together with
767{,}520 multiple-choice inferences over a stratified SuperGPQA sample and
GPQA Diamond, every individual vote is released openly, with abstentions
retained as first-class outcomes --- the joint answer distribution that
public leaderboards discard. Finally, we verify the law experimentally,
including on data never used in its calibration: conversion rates fitted once
on SuperGPQA at a 40:60 vote split transfer unchanged to GPQA Diamond and the
forensic benchmark, and the derived metrics are compared head-to-head as
predictors of realised lift with $R^2$ and Spearman's $\rho$. We therefore
seek to answer the following research questions in this paper:

\begin{itemize}
  \item \textbf{RQ1}: To what extent does a parameter-free swap law predict the
        lift of a weighted two-model ensemble, and does it capture the
        underlying mechanism rather than a mere correlate?
  \item \textbf{RQ2}: How do the accuracy gap and collective accuracy of a
        model pair govern when weighted pooling helps or harms, and what vote
        weight does each regime demand?
  \item \textbf{RQ3}: Do conversion rates calibrated once on one benchmark
        transfer to other benchmarks, including agentic tool-use tasks that
        differ in format, difficulty and dependence structure?
\end{itemize}

The law is the primary contribution; the corpus and the novel agentic
benchmark are what make its verification possible. We release the corpus as
\emph{deciban}: the vote-level records, grades, gold answers and
abstention and error reasons underlying every result in this paper
(Appendix~\ref{app:data}).

\section{Related Work}\label{sec:related}

The claim that imperfect voters can jointly beat their best member is the
Condorcet Jury Theorem, whose modern treatments established the sharp
transition in majority correctness as independent voters' competence passes
one half~\cite{grofman1983thirteen,boland1989majority}, with
heterogeneous-competence
generalisations~\cite{owen1989proving,kirstein2023generalized}. Its
guarantee erodes under dependent
ballots~\cite{ladha1992condorcet,berg1993condorcets,boland1989modelling},
and its premises have been argued jointly
unsatisfiable~\cite{dietrich2008premises}. Weighting ballots by competence
is a
standard ensemble primitive~\cite{dietterich2000ensemble,kuncheva2012weighted}
with online mistake bounds~\cite{littlestone1994weighted}; what is not
standard is
deriving the weight from a structural identity rather than fixing or learning
it. Under $K$-way plurality with symmetric distractors the analogous
competence threshold is $1/K$ rather than one half
(Section~\ref{sec:simulation}).

A parallel tradition quantifies \emph{why} combination helps: the ambiguity
decomposition~\cite{krogh1995neural}, its bias--variance--covariance
counterpart~\cite{ueda1996generalization}, a recent unified account proving
diversity is coupled to bias and variance rather than
free-standing~\cite{wood2023unified}, and its hard extension to zero-one
loss~\cite{jiang2017generalized}. Pairwise measures are built on the
2$\times$2 correct/incorrect table~\cite{kuncheva2003measures}; only
complementary errors can be repaired~\cite{sharkey1997combining,wang2002hybrid},
and disagreement is \emph{signed}, decomposing error by whether the
ensemble is right or wrong~\cite{brown2010good}, and diversity has
been treated as a quantity to be managed explicitly~\cite{brown2005managing}.
The recurring empirical verdict, however, is that structural diversity
measures track realised gain weakly and unstably across datasets and
combiners~\cite{kuncheva2003measures,tang2006analysis,shipp2002relationships,tong2019diversity,gong2019diversity}
--- in part because raw measures are coupled to the marginal accuracies. This
is the prior any new indicator must beat, and it directly motivates the
accuracy adjustment of Section~\ref{sec:phiadj}. Closer to prediction,
diversity-aware models predict majority-vote accuracy for many-member
ensembles~\cite{durrant2020diversity}, focal diversity selects
high-accuracy deep ensembles~\cite{wu2021boosting}, and competence-derived
optimal voting weights are classical~\cite{nitzan1982optimal}; under
zero--one loss our rescue mass and oracle ceiling coincide with the
complementarity potentials of human--AI
teaming~\cite{hemmer2025complementarity}. What remains open is the
weight-dependent realisation of rescue and damage for two
repeated-answer-share voters.

For large language models the literature supplies the phenomena but little of
the structure.
Self-consistency showed plurality voting over sampled reasoning paths reliably
improves a single model~\cite{wang2022selfconsistency}, with accuracy scaling
in the number of sampled agents~\cite{li2024more}; distinct models are
combined by learned fusion~\cite{jiang2023llmblender}, layered
mixtures-of-agents~\cite{wang2024mixtureofagents}, multi-agent
debate~\cite{du2023improving}, cost-driven cascades~\cite{chen2023frugalgpt}
and learned routing~\cite{lu2024routing,ong2024routellm}, now surveyed as a
taxonomy of fusion strategies~\cite{mienye2025ensemble}. The meta-analytic
caution that combinations on average underperform their best
member~\cite{vaccaro2024when} --- with complementarity
conditions framed for human--AI pairings in~\cite{choudhary2023humanai} ---
is the empirical face of the damage penalty we make explicit. To the best of
our knowledge, none of this work
offers a closed-form account of \emph{when} adding a weaker model helps:
combinations are validated benchmark by benchmark, with weights chosen
empirically or learned.

Against this backdrop there is a settled discipline of estimating combination
weights out-of-sample: stacking demands out-of-fold
predictions~\cite{ting1999issues,breiman1996stacked}, in-sample weight
optimisation overstates gains~\cite{yao2018stacking}, and ensemble
selection overfits without held-out
validation~\cite{caruana2006getting,partalas2010ensemble,large2019probabilistic}.
Whether combining beats the single best member at all is
contested~\cite{deroski2004combining,bauer1999empirical}; model soups and
routing benchmarks make held-out, cross-distribution evaluation
first-class~\cite{wortsman2022model,hu2024routerbench}, and even
``held-out'' benchmarks may be
contaminated~\cite{deng2024investigating,balloccu2024leak,xu2024benchmark}.
Our deployment rule is an in-sample argmax of the kind these works
scrutinise, which is why the cross-dataset transfer of
Section~\ref{sec:validation} --- from multiple-choice calibration to an
agentic forensic benchmark of a different format --- is central rather than
incidental to our claims.

Reading these literatures together, three gaps remain. Firstly, no prior work we know of offers a closed-form map from an
interpretable complementarity indicator to the lift of a \emph{weighted
two-model} vote over repeated answer shares; raw diversity measures are
also coupled to base accuracy, part of why they correlate weakly with
gain. Secondly, the directional claims of the LLM and
human--AI evidence (complementarity helps, the gap hurts) have rarely been
tested as \emph{quantitative} predictors, with reported $R^2$ against
realised lift, across tasks of genuinely different format. Finally, the vote-level data such
a test requires --- the joint answer distribution, not leaderboard marginals
--- is almost never released. This paper supplies all three: the law, the
head-to-head predictive test spanning an agentic benchmark, and the open
corpus.

\section{A Weighted Swap Law for Model Pairs}\label{sec:formal}

We seek to derive, from first principles, an exact account of when adding a
second model to a primary model by weighted plurality voting improves
accuracy, and to extract from it a compact heuristic computed from a handful
of interpretable quantities. We proceed in five steps: definitions
(Section~\ref{sec:defs}); a complete, assumption-free decomposition of
ensemble lift (Section~\ref{sec:fullmodel}); the \emph{weighted swap
heuristic} extracted from it (Section~\ref{sec:heuristic}); an
accuracy-adjusted correctness correlation, $\phiadj$, which reduces the
rescue mass to marginal accuracies and one dependence parameter
(Section~\ref{sec:phiadj}); and an oracle-selection ceiling for the pair
(Section~\ref{sec:ceiling}).

\subsection{Definitions}\label{sec:defs}

Let $Q=\{1,\ldots,n\}$ be a set of questions. A \emph{primary} model $A$ and a
\emph{secondary} model $B$ each answer every question $N$ times. For question
$i$, let $\mathbf{s}_A(i)$ and $\mathbf{s}_B(i)$ be the models' answer-share
vectors over $K$ output categories, so that
$\sum_{j=1}^{K} s_{Aj}(i)=\sum_{j=1}^{K} s_{Bj}(i)=1$. The categories may
include a \emph{null} category recording abstention (e.g.\ an unparseable
response), which can win a vote but is never correct.

The primary model carries fixed weight $1$ and the secondary carries weight
$x\ge 0$, giving the weighted vote vector:
\begin{equation}\label{eq:vote}
\mathbf{v}_x(i)=\mathbf{s}_A(i)+x\,\mathbf{s}_B(i).
\end{equation}
Setting $x=0$ recovers the primary alone, $x=1$ gives the models equal
influence and $x\rightarrow\infty$ approaches the secondary alone (exactly so
when the secondary's plurality is unique; at any finite weight the
primary's votes break secondary ties they distinguish).

Let $y_i$ denote the correct answer to question $i$ and let
$T_x(i)=\arg\max_j v_{x,j}(i)$ be the set of categories tied for the highest
weighted vote. The ensemble's correctness score on question $i$ awards
fractional credit on ties:
\begin{equation}\label{eq:score}
P_x(i)=
\begin{cases}
1/\lvert T_x(i)\rvert, & y_i\in T_x(i),\\
0, & \text{otherwise}.
\end{cases}
\end{equation}
The weighted-plurality accuracy is $A_x=\frac{1}{n}\sum_i P_x(i)$, and writing
$\okA(i)=\mathbf{1}[A\text{ is correct on } i]$ (with $\okB$ defined
similarly), the models' standalone accuracies are $p=\Pr(A\text{ correct})$
and $q=\Pr(B\text{ correct})$. We define the primary to be the more accurate
model, $p\ge q$, and call $\Delta=p-q\ge 0$ the \emph{accuracy gap} and
$m=(p+q)/2$ the \emph{collective accuracy}. The object of study is the
\emph{lift} of the weighted system over the primary:
\begin{equation}\label{eq:lift}
L(x)=A_x-p=\frac{1}{n}\sum_{i=1}^{n}\bigl[P_x(i)-\okA(i)\bigr].
\end{equation}

Every question falls into exactly one of four sets according to the models'
standalone correctness (stated for binary correctness; on the rare
standalone ties the scores turn fractional and the sets become the fuzzy
cell weights of Section~\ref{sec:estimation}, under which the additive
identities below continue to hold; the selector and probability readings
are binary-case statements):
the \emph{rescue set} $\mathcal{R}$ ($A$
wrong, $B$ right), the \emph{damage set} $\mathcal{D}$ ($A$ right, $B$
wrong), the both-correct set $\mathcal{C}$ and the both-wrong set
$\mathcal{Z}$. Writing
$r$, $d$, $c$ and $z$ for their normalised sizes, we have $r+d+c+z=1$,
$p=c+d$ and $q=c+r$. Subtracting the latter two identities gives the
\emph{accuracy-gap identity}:
\begin{equation}\label{eq:gapidentity}
\Delta=p-q=(c+d)-(c+r)=d-r,
\qquad\text{i.e.}\qquad d=r+\Delta.
\end{equation}
Eq.~(\ref{eq:gapidentity}) is exact and assumption-free: the primary's accuracy
advantage is precisely the excess of damage opportunity over rescue
opportunity. The accuracy gap is therefore not merely correlated with the
prospects of an ensemble; it directly determines how many more questions can
be damaged than rescued.

\subsection{The Complete Lift Model}\label{sec:fullmodel}

We now decompose Eq.~(\ref{eq:lift}) exactly. Define four weight-dependent
conversion rates, each a conditional expectation of the vote outcome on one of
the four sets: the \emph{rescue conversion}
$\alpha_x=E[P_x(i)\mid i\in\mathcal{R}]$, the \emph{damage conversion}
$\gamma_x=E[1-P_x(i)\mid i\in\mathcal{D}]$, the \emph{both-wrong repair}
$\beta_x=E[P_x(i)\mid i\in\mathcal{Z}]$ and the \emph{both-right corruption}
$\kappa_x=E[1-P_x(i)\mid i\in\mathcal{C}]$. In words, $\alpha_x$ is the
rate at which pooling at weight $x$ successfully converts rescue
opportunities, whilst $\gamma_x$ is the rate at which it destroys answers
the primary already had right.

Consider the contribution $P_x(i)-\okA(i)$ of each question to
Eq.~(\ref{eq:lift}). On $\mathcal{R}$ the primary is wrong, so the
contribution is $P_x(i)$ and rescue questions contribute $\alpha_x r$ in
aggregate. On $\mathcal{D}$ the primary is right, so the contribution is
$P_x(i)-1=-(1-P_x(i))$ and damage questions contribute $-\gamma_x d$. The same
reasoning gives $\beta_x z$ on $\mathcal{Z}$ and $-\kappa_x c$ on
$\mathcal{C}$. Summing the four contributions yields the complete model:
\begin{equation}\label{eq:exact}
L(x)=\alpha_x r-\gamma_x d+\beta_x z-\kappa_x c.
\end{equation}
No fitted parameter or approximation appears in Eq.~(\ref{eq:exact}): lift is
successful rescues, less converted damage, plus both-wrong repairs, less
both-right corruption.

\subsection{The Weighted Swap Heuristic}\label{sec:heuristic}

The first two terms of Eq.~(\ref{eq:exact}) concern the questions on which the
models disagree in correctness; these are the questions weighting exists to
arbitrate. We accordingly define the \emph{weighted swap mass}:
\begin{equation}\label{eq:sstar}
\Sstar(x)=\alpha_x r-\gamma_x d,
\end{equation}
so that $L(x)=\Sstar(x)+\varepsilon(x)$ with residual
$\varepsilon(x)=\beta_x z-\kappa_x c$. The residual is small whenever the vote
rarely repairs questions both models answer incorrectly ($\beta_x z$ small)
and rarely corrupts questions both answer correctly ($\kappa_x c$ small); we
quantify both empirically in Section~\ref{sec:calibration}. When it is small,
\begin{equation}\label{eq:heuristic}
L(x)\approx \Sstar(x),
\end{equation}
which we term the \emph{weighted swap heuristic}. An optional coefficient $\hat L_\lambda(x)=\lambda\,\Sstar(x)$,
$\lambda\approx 1$, can sharpen magnitudes; every $\lambda>0$ preserves
the maximiser, so $\lambda$ never affects weight selection.

Substituting the accuracy-gap identity of Eq.~(\ref{eq:gapidentity}) into
Eq.~(\ref{eq:sstar}) gives the rescue--damage representation:
\begin{equation}\label{eq:sstargap}
\Sstar(x)=(\alpha_x-\gamma_x)\,r-\gamma_x\Delta.
\end{equation}
Eq.~(\ref{eq:sstargap}) separates lift into a net converted rescue opportunity,
$(\alpha_x-\gamma_x)r$, and an accuracy-gap penalty, $-\gamma_x\Delta$. Three
consequences follow immediately. Firstly, if $\alpha_x<\gamma_x$ then both
terms are non-positive and the swap mass cannot be positive; when the
concordant-cell residual is negligible, such a weight cannot help. Secondly,
$\alpha_x>\gamma_x$
is necessary for positive swap mass when $\Delta>0$ but not sufficient: for $r>0$
and $\gamma_x>0$, positive swap mass requires
\begin{equation}\label{eq:threshold}
\frac{\alpha_x}{\gamma_x}>1+\frac{\Delta}{r},
\end{equation}
i.e.\ the rescue-conversion advantage must be large enough to overcome the
larger damage pool. Finally, since $\Sstar(0)=0$, primary-only deployment is
always available, and the selection rule is to use the secondary only if
$\max_{x>0}\Sstar(x)>0$, choosing $\hat x=\arg\max_{x\ge 0}\Sstar(x)$ ---
a rule for realised lift to the extent that the residual is negligible
across the candidate weights, which Section~\ref{sec:calibration} verifies
across the full weight grid. $\Sstar(x)$ is piecewise constant, changing
only at finitely many breakpoint weights; we evaluate a fixed rational grid
and report grid optima.

\subsection{Accuracy-Adjusted Correlation}\label{sec:phiadj}

The rescue mass $r$ in Eq.~(\ref{eq:sstargap}) is a joint property of the two
models. We now express it through the models' marginal accuracies and a single
normalised dependence parameter. Let $X=\okA$ and $Y=\okB$ be the models'
binary correctness indicators on a uniformly selected question, so
$E[X]=p$, $E[Y]=q$ and $E[XY]=c$. The Pearson correlation of two binaries is
the $\phi$ coefficient:
\begin{equation}\label{eq:phi}
\phi=\frac{c-pq}{\sqrt{p(1-p)\,q(1-q)}}.
\end{equation}
Raw $\phi$, however, is not on a common attainable-maximum scale across
model pairs: the overlap $c$
is at most $c_{\max}=\min(p,q)=q$ (attained when every secondary-correct
question is also primary-correct), so the maximum feasible correlation
depends on the marginal accuracies themselves. Substituting $c_{\max}=q$
into Eq.~(\ref{eq:phi}):
\begin{equation}\label{eq:phimax}
\phimax=\frac{q-pq}{\sqrt{p(1-p)\,q(1-q)}}=\sqrt{\frac{q(1-p)}{p(1-q)}}.
\end{equation}
We accordingly normalise by this maximum, $\phiadj=\phi/\phimax$: the
classical $\phi/\phi_{\max}$ coefficient of the psychometric literature,
equal for $p\ge q$ to Loevinger's
$H$~\cite{loevinger1948,cureton1959,davenport1991}, which we call the
\emph{accuracy-adjusted correlation} for its role here. It measures how
close the pair sits to the maximum correctness dependence its accuracies
permit. Dividing Eq.~(\ref{eq:phi}) by
Eq.~(\ref{eq:phimax}), the square roots cancel:
\begin{equation}\label{eq:phiadjdef}
\phiadj=\frac{c-pq}{q-pq}=\frac{c-pq}{q(1-p)}.
\end{equation}
From Eq.~(\ref{eq:phiadjdef}),
$1-\phiadj=\bigl(q(1-p)-c+pq\bigr)/\bigl(q(1-p)\bigr)=(q-c)/\bigl(q(1-p)\bigr)$,
and since $q=c+r$ the numerator is exactly the rescue mass. Rearranging:
\begin{equation}\label{eq:rescue}
r=q\,(1-p)\,(1-\phiadj).
\end{equation}
Adjusted $\phi$ is therefore algebraically contained in the rescue-set size:
the rescue mass factorises into the secondary's competence $q$, the primary's
error opportunity $1-p$ and the correctness complementarity $1-\phiadj$. At
$\phiadj=1$ the secondary has no rescue role; at $\phiadj=0$ the rescue mass
equals its value under independence, $q(1-p)$; and $\phiadj<0$ indicates
models more complementary than independence predicts --- equivalently, $1-\phiadj$ is the
ratio of observed rescue mass to that expected under independence. Whilst
$\phiadj\le 1$ always, its feasible lower bound depends on the marginals:
it is an accuracy-normalised overlap measure, not a $[-1,1]$ correlation. Substituting
Eq.~(\ref{eq:rescue}) into Eq.~(\ref{eq:sstargap}) yields the structural form
of the swap law, on which the remainder of this paper rests:
\begin{equation}\label{eq:structural}
\Sstar(x)=(\alpha_x-\gamma_x)\,q(1-p)(1-\phiadj)-\gamma_x\Delta.
\end{equation}
The swap mass is secondary competence, times primary error opportunity,
times complementarity, times net rescue conversion, less the accuracy-gap
damage.

Two further readings of Eq.~(\ref{eq:structural}) are worth noting. Holding
the conversion rates fixed, its derivative in $\phiadj$ is
$-(\alpha_x-\gamma_x)\,q(1-p)$: in the useful-conversion regime
($\alpha_x>\gamma_x$) greater complementarity increases the predicted swap
mass. Its derivative
in $\Delta$ at fixed $m$ and $\phiadj$ is
$-\tfrac{1}{2}(\alpha_x-\gamma_x)(1-\Delta)(1-\phiadj)-\gamma_x$, which is
non-positive whenever $\alpha_x\ge\gamma_x$.

\subsection{The Oracle-Selection Ceiling}\label{sec:ceiling}

Finally, we bound what any combiner restricted to \emph{selecting} between
the two models' standalone answers can achieve; pooling itself can exceed
this only through the both-wrong repair term $\beta_x z$ of
Eq.~(\ref{eq:exact}) --- in the binary case $L(x)\le r+\beta_x z$, and
universally $L(x)\le 1-p$ --- and Section~\ref{sec:calibration} shows
both concordant-cell components are individually negligible in practice. An oracle selector that
chooses the correct model whenever exactly one of the two is correct is
the best case $\alpha=1$, $\gamma=0$ of Eq.~(\ref{eq:sstar}), and its lift is
exactly the rescue mass:
\begin{equation}\label{eq:oracle}
L_{\mathrm{oracle}}=r=q\,(1-p)\,(1-\phiadj).
\end{equation}
Every rescue question is both secondary-correct and primary-wrong, so
$r\le\min(q,\,1-p)$. Writing $q=m-\Delta/2$ and $1-p=1-m-\Delta/2$, both
bounds shrink linearly in the gap and we obtain the ceiling:
\begin{equation}\label{eq:ceiling}
L_{\mathrm{oracle}}\le\min(q,\,1-p)=\min(m,\,1-m)-\frac{\Delta}{2}.
\end{equation}
The ceiling is tight, and it makes the cost of an accuracy gap unavoidable: at
fixed collective accuracy, every additional point of gap removes half a point
from the maximum lift any selection mechanism --- however clever --- can
extract from the pair. The accuracy of any such selector is accordingly
bounded by $p+\min(m,1-m)-\Delta/2$; the pooled vote can exceed this only
through both-wrong repair.

\section{Accuracy Gap \& Collective Accuracy}\label{sec:simulation}

\begin{figure*}[t]
  \centering
  \includegraphics[width=\textwidth]{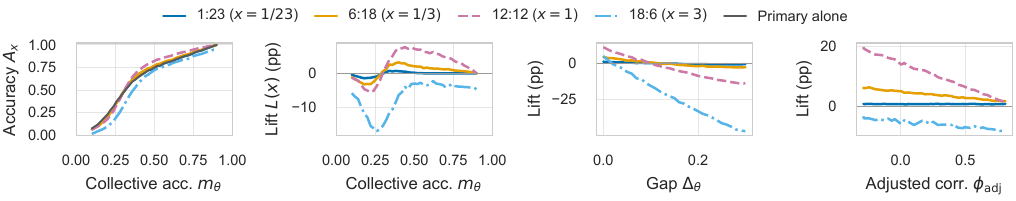}
  \caption{Monte Carlo behaviour of the finite weighted vote under
  question-clustered competence; the legend gives the secondary:primary
  split of the pooled vote. Left to right: accuracy $A_x$ vs $m_\theta$; lift
  over the primary's own plurality vs $m_\theta$ (both at
  $\Delta_\theta=0.10$); lift vs $\Delta_\theta$ ($m_\theta=0.30$);
  lift vs $\phiadj$ ($m_\theta=0.40$, $\Delta_\theta=0.10$). Panels
  1--3 fix the latent joint-knowledge overlap (measured
  $\phiadj\approx 0.5$ mid-range)}
  \Description{Four Monte Carlo panels: accuracy and lift against collective accuracy, lift against accuracy gap, and lift against adjusted correlation, for five vote splits.}
  \label{fig:theory}
\end{figure*}

We seek to demonstrate how collective accuracy, the accuracy gap and the
correlation of errors govern the behaviour of the finite weighted vote,
before turning to empirical data. We use a deliberately simple Monte Carlo
illustration: $K=4$ options, $N=24$ samples per model and fractional tie
credit as in Eq.~(\ref{eq:score}), with competence clustered by question
--- each question is either \emph{known} to a model, carrying per-sample
probability $0.95$ of the correct option, or \emph{unknown} ($0.05$, the
remainder spread evenly over the distractors). Both levels sit far from
the plurality threshold $\theta=1/K$, so these votes are nearly
deterministic. A $45\%$ share of questions is instead \emph{contested}: neither model
knows them, both sample at their mean rates, and the finite vote
decides. Contested questions are where pooling can hurt --- a
weaker model's scattered votes can flip a narrow plurality --- and they
drive every negative region of Figure~\ref{fig:theory}. The joint-knowledge probability sets the dependence: panels 1--3 fix the
clustered share's latent overlap, giving measured $\phiadj\approx 0.5$
mid-range (it varies with the marginals along each sweep), and panel 4
sweeps it. We
sweep the per-sample analogues $m_\theta=(\theta_A+\theta_B)/2$ and
$\Delta_\theta=\theta_A-\theta_B$ of the plurality-level $m$ and
$\Delta$ of Section~\ref{sec:formal}, where $\theta$ is a model's mean
per-sample accuracy; lift is measured, per Eq.~(\ref{eq:lift}), against
the primary's own 24-vote plurality accuracy $A_0$. Each point averages
15{,}000 trials; legend ratios split the pooled vote secondary:primary
out of 24 ($x=2/3$, the 40:60 split of Section~\ref{sec:calibration},
sits between $6{:}18$ and $12{:}12$).

As shown in Figure~\ref{fig:theory} (left), accuracy follows a soft
S-shape --- contested questions give the vote a knee near the plurality
threshold --- then climbs towards the pair's joint knowledge. Lift
over the primary's own plurality (second panel) is hump-shaped: the rescue
opportunity $q(1-p)$ of Eq.~(\ref{eq:rescue}) is scarce whilst the
secondary knows little and vanishes as the primary's knowledge saturates,
so the balanced $12{:}12$ vote peaks at ${\approx}{+}8$pp near
$m_\theta=0.45$ ($6{:}18$ at ${\approx}{+}3$pp) --- yet every weight is
harmful below $m_\theta\approx 0.3$, where contested questions dominate
the ledger and rescue is scarce ($-5.5$pp at $12{:}12$). The extremes fail in opposite ways: $1{:}23$ is
too small to overturn the primary anywhere and captures almost none of
the available rescue, whilst $18{:}6$ hands control to the weaker model
and is harmful at every $m_\theta$. (Figure~\ref{fig:decomp},
Appendix~\ref{app:weights}, separates the self-consistency and diversity
contributions.)

The third panel sweeps the per-sample gap at fixed $m_\theta=0.30$.
Between equals even heavy over-weighting costs little ($18{:}6$ gains
$+4.9$pp at $\Delta_\theta=0$), however widening the gap lowers lift at
every weight, qualitatively as the penalty $-\gamma_x\Delta$ of
Eq.~(\ref{eq:structural}) requires, and the collapse is weight-ordered:
$18{:}6$ turns harmful by $\Delta_\theta\approx 0.02$ and falls to
$-47$pp, whilst by $\Delta_\theta\approx 0.12$ even the $6{:}18$ and
balanced votes turn negative ($-14$pp at $12{:}12$ by
$\Delta_\theta=0.30$) --- the forensic regime of
Section~\ref{sec:cyberlift}. Greater secondary weight
tolerates a smaller accuracy gap.

The fourth panel varies the dependence itself: lift falls essentially
linearly as $\phiadj$ rises --- the $(1-\phiadj)$ release of rescue mass
that Eq.~(\ref{eq:structural}) prescribes --- from $+20$pp at
$\phiadj=-0.3$ for the balanced vote (negative $\phiadj$: more
complementary than independence, Section~\ref{sec:phiadj}), through $+7$pp at
$\phiadj=0.5$, to ${\approx}{+}1$pp at its ceiling ($\phiadj\approx
0.8$), where the clustered knowledge nests and only contested questions
keep rescue open. The $18{:}6$ vote is harmful at
every dependence level here: weight, not complementarity, is its problem.
Complementarity, not raw accuracy, is the fuel of lift: high
overall accuracy does not preclude it. These simulations fix the
qualitative mechanism we test next.

\section{Datasets \& Experimental Setup}\label{sec:datasets}

We seek to test the swap law on heterogeneous tasks: two multiple-choice
science benchmarks and one agentic digital-forensics benchmark, summarised in
Table~\ref{tab:datasets}. All three use the same fleet of ten open-weight
models (20--31B parameters), the same 24 samples per model--question pair and
temperature 1.0 throughout, with chain-of-thought reasoning enabled. Answer
positions on both multiple-choice question (MCQ) datasets were randomised
per sample with a balanced
schedule shared across models; for 4-option questions the 24 samples cover
all 24 orderings exactly once.

\begin{table}[h]
\centering
\renewcommand{\arraystretch}{1.15}
\caption{The three evaluation datasets.}
\label{tab:datasets}
\small
\begin{tabular}{|l|r|l|r|}
\hline
\textbf{Dataset} & \textbf{Questions} & \textbf{Answer format} & \textbf{Trials} \\ \hline
SuperGPQA (sample) & 3,000 & MCQ, 4--10 options & 720{,}000 \\ \hline
GPQA Diamond & 198 & MCQ, 4 options & 47{,}520 \\ \hline
Forensic (agentic) & 98 & free text, tool use & 23{,}520 \\ \hline
\end{tabular}
\\[3pt]
{\footnotesize Trials = graded model--question--sample outcomes across the fleet, abstentions included.}
\end{table}

\textbf{SuperGPQA.} A fixed 3{,}000-question stratified sample of the
26{,}529-question SuperGPQA benchmark~\cite{du2025supergpqa} (4--10 options
per question), drawn once before any inference with proportional allocation
over discipline--difficulty strata plus a stress phase oversampling hard
strata; it serves as the calibration set, and all SuperGPQA statistics
target this fixed design rather than the full benchmark.

\textbf{GPQA Diamond.} All 198 questions of the graduate-level GPQA Diamond
benchmark~\cite{rein2024gpqa} (4 options).

\textbf{Forensic.} Five real digital-forensics challenges (98 questions):
disk-image and network-capture analysis performed agentically --- multi-turn
tool use inside a network-isolated sandbox --- with free-text answers graded
offline against gold answers, using an external judge model for rubric-tier
questions. Answers are unconstrained strings; vote categories map every
answer graded correct to one gold category, group incorrect answers by
normalised string equality and treat unanswered questions as abstentions.

\begin{table}[h]
\centering
\renewcommand{\arraystretch}{1.15}
\caption{Model accuracy by dataset: single-sample accuracy and 24-vote plurality accuracy (\%).}
\label{tab:model_accuracy}
\scriptsize
\begin{tabular}{|l|r|r|r|r|r|r|}
\hline
\textbf{Model} & \textbf{SG-1} & \textbf{SG-24} & \textbf{GD-1} & \textbf{GD-24} & \textbf{F-1} & \textbf{F-24} \\ \hline
Gemma-4-31B & 66.1 & 69.0 & 86.0 & 88.1 & 20.2 & 22.4 \\ \hline
Qwen3.6-27B & 64.9 & 67.9 & 84.4 & 86.1 & 26.5 & 31.1 \\ \hline
Gemma-4-26B-A4B & 61.0 & 64.8 & 80.3 & 84.1 & 15.3 & 13.8 \\ \hline
Qwen3-30B-Think. & 56.0 & 59.4 & 71.4 & 72.2 & 10.8 & 13.3 \\ \hline
DiffusionGemma-26B & 49.5 & 54.9 & 66.2 & 71.5 & 0.8 & 0.0 \\ \hline
GLM-4.7-Flash & 50.3 & 53.2 & 63.6 & 66.2 & 9.2 & 12.8 \\ \hline
Nemotron-3-Nano & 47.4 & 50.7 & 63.7 & 65.7 & 12.1 & 13.8 \\ \hline
GPT-OSS-20B & 44.4 & 49.5 & 65.0 & 69.4 & 1.1 & 0.0 \\ \hline
Magistral-Small & 36.5 & 47.4 & 48.4 & 58.6 & 6.8 & 11.2 \\ \hline
MedGemma-27B & 39.7 & 46.7 & 48.3 & 53.5 & 5.4 & 7.1 \\ \hline
\end{tabular}
\\[3pt]
{\footnotesize SG = SuperGPQA, GD = GPQA Diamond, F = Forensic; -1 = single sample, -24 = plurality of 24 samples (fractional tie credit). Unparseable outputs count as abstentions.}
\end{table}

Table~\ref{tab:model_accuracy} reports single-sample and 24-vote plurality
accuracy. The fleet spans a wide range on every dataset (46.7\%--69.0\%
plurality on SuperGPQA; 0\%--31.1\% forensic), so pairs realise accuracy
gaps from under 1pp to over 30pp; the forensic tasks are much harder, with
the strongest model at 26.5\% per sample and two models never winning a
plurality.

Abstention is a first-class output in our framework (the null category of
Section~\ref{sec:defs}): on the MCQ corpora unparseable inferences ---
dominated by reasoning truncated at the token cap --- enter the vote vectors
as null votes rather than guesses; on the forensic tasks the equivalent
is a question left unanswered at trace termination. Per-model abstention rates
are reported in Table~\ref{tab:abstentions} (Appendix~\ref{app:pairs}).

\subsection{Estimation}\label{sec:estimation}

All quantities of Section~\ref{sec:formal} are estimated per ordered model
pair, per dataset. For each question we form the empirical share vectors of
Eq.~(\ref{eq:vote}) from the 24 votes (canonical answer categories plus null)
and score the pooled vote by Eq.~(\ref{eq:score}). A model's own correctness
$\okA(i)$ is its 24-vote plurality score at $x=0$, which is fractional on
ties; the set masses $r,d,c,z$ are then estimated as means of products (e.g.\
$r=\frac{1}{n}\sum_i(1-\okA(i))\,\okB(i)$), under which the identities of
Eq.~(\ref{eq:gapidentity}) and Eq.~(\ref{eq:rescue}) continue to hold exactly.
Conversion rates generalise correspondingly. Writing $a_i=\okA(i)$,
$b_i=\okB(i)$ and the fuzzy weights $w_i^{\mathcal R}=(1-a_i)b_i$,
$w_i^{\mathcal D}=a_i(1-b_i)$, $w_i^{\mathcal Z}=(1-a_i)(1-b_i)$,
$w_i^{\mathcal C}=a_ib_i$:
\begin{equation}\label{eq:fuzzyrates}
\alpha_x,\beta_x=\frac{\sum_i(P_x(i)-a_i)\,w_i^{\bullet}}{\sum_i w_i^{\bullet}},
\qquad
\gamma_x,\kappa_x=\frac{\sum_i(a_i-P_x(i))\,w_i^{\bullet}}{\sum_i w_i^{\bullet}},
\end{equation}
with $\bullet$ the corresponding cell ($\mathcal R,\mathcal Z$
gain-signed; $\mathcal D,\mathcal C$ loss-signed); the decomposition of
Eq.~(\ref{eq:exact}) remains exact, and each rate reduces to its
conditional reading ($\alpha_x=E[P_x\mid\mathcal{R}]$, etc.) in the
binary case. With fractional scores the rates are signed contributions and
the sign-based consequences of Section~\ref{sec:heuristic} are stated for
the binary case; ties are rare (2.65\% of model--question votes), and the
Bernoulli-form denominator of Eq.~(\ref{eq:phi}) is retained as the
definition of empirical $\phi$.
Cells with zero mass have undefined rates and contribute zero. Within each
pair the primary is the model with the higher plurality accuracy, giving 45
ordered pairs per dataset. Weights are evaluated on a fixed 25-point
rational grid from $x=1/24$ to $x=24$ ($x=a/b$ with the vote computed as
$b\,\mathbf{s}_A+a\,\mathbf{s}_B$ in integer counts), so tie detection is
exact; where several weights tie on $\Sstar$ we take the smallest.

\begin{figure}[h]
  \centering
  \includegraphics[width=\columnwidth]{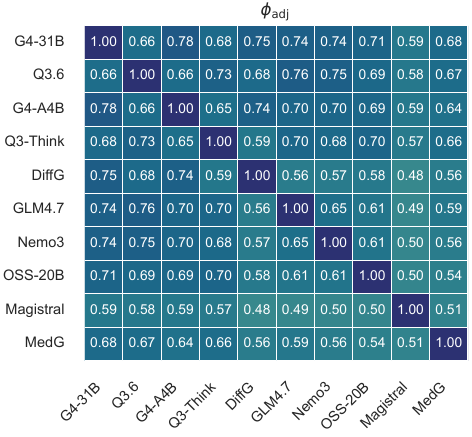}
  \caption{Accuracy-adjusted correlation $\phiadj$ across model pairs, SuperGPQA}
  \Description{Heatmap of accuracy-adjusted correlation for all SuperGPQA model pairs.}
  \label{fig:phiadj_sg}
\end{figure}

Figure~\ref{fig:phiadj_sg} charts $\phiadj$ across the SuperGPQA pairs
(models ordered by accuracy; the raw $\phi$ counterpart is
Figure~\ref{fig:phi_sg} in Appendix~\ref{app:weights}, and both quantities
are computable per pair for every dataset from the released votes). The
adjustment matters in
practice: raw $\phi$ spans 0.37--0.71 on SuperGPQA and is systematically
depressed for pairs with large accuracy gaps, precisely because $\phimax<1$;
$\phiadj$ removes this artefact and spans 0.48--0.78. On the forensic tasks
dependence is more extreme in both directions ($\phiadj$ from 0.26 to 1.00)
--- some pairs are as correlated as their accuracies permit, leaving no
rescue mass at all.

\section{Calibrating the Heuristic at 40:60}\label{sec:calibration}

We seek to calibrate the weighted swap heuristic once, on the largest
dataset, and then test whether it transfers unchanged to the others. The
operating weight is fixed throughout at a secondary:primary 40:60 vote
split ($x=2/3$):
Section~\ref{sec:simulation} shows sub-balanced pooling the most
gap-tolerant regime, and $x=2/3$ is the largest grid weight at which
SuperGPQA's mean pair lift remains positive (Figure~\ref{fig:meanlift},
Appendix~\ref{app:weights}).

On the 45 SuperGPQA pairs at 40:60 the measured conversion rates average
$\bar\alpha=0.338$ and $\bar\gamma=0.165$ (question-level bootstrap 95\%
intervals $[0.32,0.36]$ and $[0.16,0.17]$): the
pooled vote converts roughly a third of rescue opportunities whilst destroying
a sixth of the damage pool. Regressing observed lift on the measured swap mass
through the origin gives $\lambda=1.04$ (bootstrap interval
$[1.00,1.08]$, containing $\lambda=1$); the concordant-cell residual
$\varepsilon$ is small and slightly positive (mean $+0.18$pp against a lift
standard deviation of 1.3pp); its components are small on average
(the mean of each, $E[\beta_x z]$ and $E[\kappa_x c]$, is below $0.12$pp
in magnitude on every dataset), and across the full weight grid $|\varepsilon|$ averages
$\le 0.2$pp (maximum 1.64pp over all 3{,}510 pair--weight cells). Selecting
each pair's weight in-sample by the swap-mass rule recovers a lift within
0.1pp of the grid optimum for 131 of 135 pairs (mean shortfall
$\le 0.02$pp; held-out, Section~\ref{sec:heldout}).

The calibrated, transportable form of the heuristic instantiates
Eq.~(\ref{eq:structural}) with the SuperGPQA-averaged conversion rates:
\begin{equation}\label{eq:shat}
\hat S=(\bar\alpha-\bar\gamma)\,q(1-p)(1-\phiadj)-\bar\gamma\,\Delta,
\end{equation}
which requires only the pair's marginal accuracies and $\phiadj$ --- no
pooled inference, though it does require labelled repeated outputs from
both models on the target task (at the boundaries $q=0$ and $p=1$, where
$\phiadj$ is undefined, $\hat S$ is computed through the equivalent form
$(\bar\alpha-\bar\gamma)r-\bar\gamma\Delta$). Throughout, $R^2$ is the squared Pearson correlation between predictor and
observed lift per dataset; for $\hat S$, which predicts lift's own scale,
we also report the identity prediction's direct error.
As shown in Figure~\ref{fig:calibration}
(Appendix~\ref{app:weights}), $\hat S$
explains the SuperGPQA lifts well ($R^2=0.71$, Spearman's $\rho=0.84$,
N=45, RMSE 0.9pp).

\begin{figure}[h]
  \centering
  \includegraphics[width=\columnwidth]{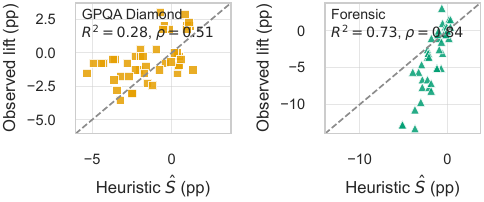}
  \caption{The SuperGPQA-calibrated heuristic transferred to GPQA Diamond and the forensic dataset}
  \Description{Scatter plots of calibrated heuristic against observed lift on GPQA Diamond and the forensic dataset.}
  \label{fig:validation}
\end{figure}

\subsection{The Heuristic Holds Across Datasets}\label{sec:validation}

We then apply Eq.~(\ref{eq:shat}) --- $\bar\alpha$, $\bar\gamma$ frozen at
their SuperGPQA values --- to the 45 GPQA Diamond and 45 forensic pairs
(Figure~\ref{fig:validation}), comparing against the single-metric
alternatives $m$, $\Delta$, $\phi$ and $\phiadj$;
Table~\ref{tab:predictors} reports $R^2$ and Spearman's $\rho$ against
observed lift.

\begin{table*}[h]
\centering
\renewcommand{\arraystretch}{1.15}
\caption{Predicting pair-level lift at the 40:60 weighting: $R^2$ and Spearman's $\rho$ against observed lift, for the swap-law quantities and single-metric alternatives.}
\label{tab:predictors}
\small
\begin{tabular}{|l|r|r|r|r|r|r|}
\hline
 & \multicolumn{2}{c|}{\textbf{SuperGPQA}} & \multicolumn{2}{c|}{\textbf{GPQA Diamond}} & \multicolumn{2}{c|}{\textbf{Forensic}} \\ \hline
\textbf{Predictor} & \textbf{$R^2$} & \textbf{$\rho$} & \textbf{$R^2$} & \textbf{$\rho$} & \textbf{$R^2$} & \textbf{$\rho$} \\ \hline
Measured swap mass $S^{\star}$ & 0.99 & 0.99 & 0.96 & 0.98 & 1.00 & 1.00 \\ \hline
Calibrated heuristic $\hat S$ & 0.71 & 0.84 & 0.28 & 0.51 & 0.73 & 0.84 \\ \hline
Rescue mass $r$ & 0.81 & 0.92 & 0.32 & 0.58 & 0.18 & 0.60 \\ \hline
Collective accuracy $m$ & 0.28 & -0.60 & 0.18 & -0.43 & 0.00 & 0.07 \\ \hline
Accuracy gap $\Delta$ & 0.56 & -0.77 & 0.18 & -0.42 & 0.68 & -0.80 \\ \hline
Adjusted correlation $\phi_{\mathrm{adj}}$ & 0.67 & -0.85 & 0.18 & -0.40 & 0.08 & -0.42 \\ \hline
Raw correlation $\phi$ & 0.00 & -0.04 & 0.01 & 0.11 & 0.08 & 0.40 \\ \hline
\end{tabular}
\\[3pt]
{\footnotesize N=45 pairs per dataset ($\phi$ and $\phi_{\mathrm{adj}}$ on the forensic dataset over the 28 pairs where defined). $\hat S$ uses conversion rates calibrated once on SuperGPQA. $S^{\star}$ requires the pooled votes (retrospective); all other rows are computable pre-pooling.}
\end{table*}

\begin{figure*}[t]
  \centering
  \includegraphics[width=0.75\textwidth]{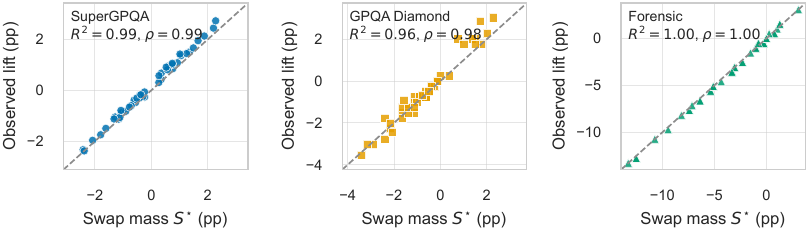}
  \caption{Measured (retrospective) swap mass against observed lift at
  40:60 on all three datasets}
  \Description{Scatter plots of measured swap mass against observed lift on all three datasets, tightly following the identity line.}
  \label{fig:exactness}
\end{figure*}

Three results stand out. Firstly, the measured swap mass $\Sstar$ --- an
identity check computed from the same pooled votes as the lift it tracks,
not a prediction --- is near-exact everywhere ($R^2\ge 0.96$ and $\rho\ge 0.98$
on all three datasets; Figure~\ref{fig:exactness}): the concordant-cell
residual of Eq.~(\ref{eq:exact}) is negligible in practice, so the
heuristic's two-term truncation loses almost nothing.

Secondly, no single quantity is stable across all three datasets. The
rescue mass leads on the MCQ datasets ($\rho=0.92$ and $0.58$) yet
degrades sharply on the forensic transfer ($R^2=0.18$), where wide gaps
make the damage term decisive; the accuracy gap shows the mirror pattern
($\rho=-0.77$ and $-0.80$, but $-0.42$ on GPQA Diamond); $\phiadj$ is
strong on SuperGPQA ($\rho=-0.85$) yet unstable elsewhere; raw $\phi$ has
little predictive power anywhere ($R^2\le 0.09$) --- without the accuracy adjustment of
Eq.~(\ref{eq:phiadjdef}), inter-model correlation says almost nothing about
ensemble prospects.

Finally, the transported heuristic $\hat S$ --- combining these
quantities --- is the most stable pre-pooling score across the three
datasets (worst-case $R^2=0.28$ against $0.18$ for the rescue mass) and
the strongest on the forensic transfer ($\rho=0.84$; question-level
bootstrap 95\% intervals $[0.26,0.65]$ on GPQA Diamond and $[0.68,0.88]$
on forensic), despite its conversion rates never being refitted. On the calibration set $|\rho|$ for $\phiadj$ is marginally higher (0.85
against 0.84); $\phiadj$ alone degrades off-calibration whilst the
heuristic does not. The forensic transfer is notable: task format (free-text, tool-mediated,
abstention-heavy), accuracy regime and dependence structure all differ,
yet the association holds (identity RMSE 3.5pp). GPQA Diamond is the weakest transfer ($R^2=0.28$): with 198 questions
per-pair lift is noisy and magnitude calibration trails a zero-lift
baseline (identity RMSE 1.9pp against 1.7pp), so off-calibration $\hat S$
is a rank screen rather than a point predictor; the ordering ($\rho=0.51$)
remains informative.

\subsection{Held-Out Validation}\label{sec:heldout}

We seek to test the swap-mass rule and the heuristic out of sample on the
calibration dataset itself: the 3{,}000 SuperGPQA questions are split at
random into a 40\% calibration set and a disjoint 60\% evaluation set, over
500 splits, at equal weighting ($x=1$), where pooling harms the average
pair (Figure~\ref{fig:meanlift}). Table~\ref{tab:heldout} reports the
results. Both transfer: weights chosen by the swap-mass rule on the
calibration questions turn a mean held-out lift of $-0.77$pp under equal
weighting into $+0.51$pp, and the heuristic calibrated on the same split
predicts held-out lift with $\rho=0.83$ whilst raw $\phi$ remains at
$\rho\le 0.03$.

\begin{table}[H]
\centering
\renewcommand{\arraystretch}{1.05}
\caption{Held-out validation on SuperGPQA at equal weighting.}
\label{tab:heldout}
\footnotesize
\begin{tabular}{|l|r|}
\hline
\textbf{Quantity} & \textbf{Value} \\ \hline
Held-out lift, swap-mass-selected weight & $+0.51$pp \\ \hline
Held-out lift, equal weighting throughout & $-0.77$pp \\ \hline
$\rho(\hat S,\ \text{held-out lift})$, coefficients transported & 0.83 \\ \hline
$\rho(\hat S,\ \text{held-out lift})$, all quantities out of sample & 0.80 \\ \hline
$\rho(\phi,\ \text{held-out lift})$, out of sample & 0.03 \\ \hline
$\bar\alpha$ (calibration split) & $0.47 \pm 0.01$ \\ \hline
$\bar\gamma$ (calibration split) & $0.28 \pm 0.01$ \\ \hline
\end{tabular}
\\[3pt]
{\footnotesize N=45 pairs; 500 random 40/60 question splits of the 3,000 SuperGPQA questions; $x=1$; means ($\pm$ sd) across splits; $\rho$ is Spearman's.}
\end{table}

\subsection{Sample Efficiency}\label{sec:samples}

We seek to compare the heuristic with estimating lift directly at equal
sample cost, since both require labelled outputs from the two models: for
$k$ of the 24 samples drawn at random per model--question cell, each is
computed on the 45 SuperGPQA pairs and ranked against the true 24-sample
lift at 40:60. Table~\ref{tab:samples} reports the results. Where repeated
samples exist direct estimation dominates ($\rho=0.90$ from two samples per
model against $0.73$ for $\hat S$), however the heuristic remains
informative from a single sample ($\rho=0.73$), where a sub-balanced vote
is degenerate (the primary's single vote cannot be outweighed) and lift
cannot be estimated at all: it screens on the per-question correctness that
ordinary single-pass evaluations already provide.

\begin{table}[H]
\centering
\renewcommand{\arraystretch}{1.05}
\setlength{\tabcolsep}{4pt}
\caption{Ranking pairs by lift from $k$ samples per model, SuperGPQA at 40:60.}
\label{tab:samples}
\footnotesize
\begin{tabular}{|l|r|r|r|r|}
\hline
\textbf{$k$} & \textbf{Labelled outputs} & \textbf{$\rho$, $\hat S$} & \textbf{$\rho$, direct} & \textbf{$\rho$, $\phi$} \\ \hline
1 & 6,000 & 0.73 & --- & $-0.27$ \\ \hline
2 & 12,000 & 0.73 & 0.90 & $-0.27$ \\ \hline
4 & 24,000 & 0.78 & 0.93 & $-0.17$ \\ \hline
8 & 48,000 & 0.80 & 0.95 & $-0.11$ \\ \hline
24 & 144,000 & 0.84 & 1.00 & $-0.04$ \\ \hline
\end{tabular}
\\[3pt]
{\footnotesize N=45 pairs; Spearman's $\rho$ against the true 24-sample lift, mean over 200 random draws of $k$ samples per model--question cell ($k=24$ is the full data). Labelled outputs $=2k\times 3{,}000$ per pair. ---: with one sample per model the 40:60 vote reduces to the primary alone (lift identically zero).}
\end{table}

\section{Ensemble Lift in Practice}\label{sec:lift}

We seek finally to quantify the lift that weighted pairing actually delivers,
and how the deployment weight must adapt across regimes (weight-response
curves are charted in Appendix~\ref{app:weights}).

On SuperGPQA, pooling at 40:60 is worthwhile for well-chosen pairs: 20 of 45
pairs gain, with the largest reaching +2.7pp
(Table~\ref{tab:top_pairs}, Appendix~\ref{app:pairs}); the top-accuracy
pair (Gemma-4-31B with Qwen3.6-27B, the two most accurate)
gains +1.5pp at 40:60 and peaks at +1.9pp at equal weighting, lifting
plurality accuracy from 69.0\% to 70.8\%. The top of
Table~\ref{tab:top_pairs} is not the most accurate pairs but the most
complementary ones amongst near-peers --- the $q(1-p)(1-\phiadj)$
geometry of Eq.~(\ref{eq:rescue}).

The structural role of the accuracy gap is visible across the full pair
population: observed lift at 40:60 declines with $\Delta$ on every dataset
($\rho=-0.77$, $-0.42$, $-0.80$), steepest on the forensic tasks where
damage conversion is largest (Figure~\ref{fig:liftdelta},
Appendix~\ref{app:weights}).

On GPQA Diamond the picture is more cautionary: only 11 of 45 pairs gain
at 40:60, and for the top-accuracy pair every positive weight is harmful
---
one of the four (of 135) where the in-sample rule misses: it selects
$x=5/4$ for a predicted $+0.13$pp that realises $-0.25$pp instead of
$\hat x=0$.
For 11 of 45 GPQA Diamond pairs (and 35 of 45 forensic pairs) no grid weight
produces positive lift: primary-only deployment is a live option, not a
failure mode.

\subsection{Lift on the Forensic Dataset}\label{sec:cyberlift}

The forensic dataset operates in a different regime: collective accuracy is
low (the harmful low-$m_\theta$ foot of Figure~\ref{fig:theory}),
accuracy gaps between
neighbouring models are large relative to rescue mass, and abstention is
heavy. Pooling at 40:60 is accordingly harmful --- only 7 of 45 pairs gain, and
the mean lift across pairs is $-3.5$pp --- however this need not make the
secondary useless; the weight regime matters.

\begin{table}[h]
\centering
\renewcommand{\arraystretch}{1.15}
\caption{Forensic dataset: weighted-plurality accuracy of the top-accuracy pair (Qwen3.6-27B primary, Gemma-4-31B secondary) as the secondary's vote weight $x$ varies.}
\label{tab:cyber_weights}
\footnotesize
\begin{tabular}{|l|r|r|r|}
\hline
\textbf{Weight $x$} & \textbf{Accuracy \%} & \textbf{Lift (pp)} & \textbf{$S^{\star}$ (pp)} \\ \hline
0 (primary only) & 31.1 & +0.00 & +0.00 \\ \hline
1/12 & 32.7 & +1.53 & +1.28 \\ \hline
1/11 & 32.7 & +1.53 & +1.28 \\ \hline
1/6 & 32.7 & +1.53 & +1.28 \\ \hline
2/3 & 30.6 & -0.51 & -0.77 \\ \hline
1 & 30.6 & -0.51 & -0.77 \\ \hline
4 & 22.4 & -8.67 & -8.93 \\ \hline
\end{tabular}
\\[3pt]
{\footnotesize 98 questions; fractional tie credit. The optimum sits on the plateau from $x=1/12$ to $1/6$.}
\end{table}

Table~\ref{tab:cyber_weights} traces the top-accuracy pair (Qwen3.6-27B primary at
31.1\%, Gemma-4-31B secondary at 22.4\%) across vote weights;
Figure~\ref{fig:cyber_weights} (Appendix~\ref{app:weights}) charts the
strongest pairs. At 40:60 the pair's lift
is $-0.5$pp, whilst de-weighting the secondary to the plateau between
$x=1/12$ and $x=1/6$ converts the same pair to +1.5pp (32.7\% against the
primary's 31.1\%). The gain rests on
two rescued questions from one challenge (one full, one fractional); a
question-level interval reaches zero. Past equal weighting the pair collapses towards secondary-only
performance ($-8.7$pp at $x\ge 4$) --- the $-\Delta$ limit of
Section~\ref{sec:heuristic} made visible. The mechanism is the one Eq.~(\ref{eq:threshold}) formalises: with
$r=2.6$pp and $\Delta=8.7$pp, weights with $\gamma_x>0$ need
$\alpha_x/\gamma_x>1+\Delta/r\approx 4.4$; on the plateau $\gamma_x=0$
and $\Sstar=\alpha_x r>0$ directly --- weights too small to outvote a
unified primary, yet large enough to break its ties.

\section{Beyond Pairs: Three-Model Ensembles}\label{sec:trios}

We close by indicating where the framework leads beyond model pairs. The
four-set partition of Section~\ref{sec:defs} generalises directly: with two
secondaries the questions fall into $2^3$ correctness cells, the accuracy-gap
identity applies to each secondary separately and the oracle-selection
ceiling of Section~\ref{sec:ceiling} becomes the \emph{union} rescue
mass. As a first empirical indication, we evaluated every trio (primary = most
accurate member) over a rational weight grid on both secondaries, on all
three datasets.

Table~\ref{tab:trios} (Appendix~\ref{app:pairs}) compares the best grid-searched trio against the best
grid-searched pair per dataset. A third member materially raises the ceiling
(union rescue mass 17.2pp against 10.9pp for the best SuperGPQA
composition), and realised lift rises on every
dataset (+3.7pp against +2.9pp on SuperGPQA; +4.9 against +4.4 on GPQA Diamond;
+3.6 against +3.1 forensic, the last converting its entire ceiling) --- though these are in-sample grid maxima with the attendant selection
optimism; best pair and trio need not share a primary, so these are not
pure third-member effects, and maximising lift favours weak primaries: the
compositions sit below the strongest single model in absolute accuracy. For the strongest primaries,
however, the third member adds almost nothing at these weights: amongst the
leading models its correct votes largely duplicate the second's, and the
gains concentrate amongst mid-accuracy trios near the plurality threshold
of Section~\ref{sec:simulation} (measured against weaker primaries).

Future work may develop the swap law fully for larger panels ---
per-member rates over the $2^k$ cells, a generalised $\phiadj$.

\section{Analysis \& Limitations}\label{sec:analysis}

We note four limitations.
Firstly, the forensic dataset has 98 questions nested in five challenges,
so per-pair lift is noisy and question-level bootstrap intervals
understate clustering; we rest on rank-level agreement ($\rho=0.84$), not
point estimates. Secondly, the 45
pairs share ten models and a common question set, so pair-level statistics
are not independent replicates; transfer is task transfer for this fixed
fleet, not model-level external validation.
Thirdly, our free-text category construction maps every correct answer to
one gold category whilst wrong answers scatter --- a normalisation unavailable at
deployment, so the forensic results are an oracle-normalised reading of
free-text pooling and per-pair weight optima are sensitive to that
construction; the MCQ datasets carry no such asymmetry. Finally, the 131/135 figure is
in-sample curve reconstruction; the question-disjoint test of
Section~\ref{sec:heldout} finds the swap-mass rule carries $+0.51$pp to
held-out questions against $-0.77$pp for equal weighting, whilst
cross-dataset transfer validates the fixed 40:60 predictor only.

\section{Conclusion}\label{sec:conclusion}

In this paper we have derived a predictive law for the uplift diversity
of thought provides in weighted LLM ensembles, verified on an open
vote-level corpus (\emph{deciban}, Appendix~\ref{app:data}) of 767{,}520
MCQ inferences and an agentic forensic benchmark.

On \textbf{RQ1}, the measured swap mass accounts for realised lift almost
exactly ($R^2\ge 0.96$, $\rho\ge 0.98$ throughout) across MCQ science
and free-text agentic forensics.
On \textbf{RQ2}, the gap governs the regime: balanced pooling rewards
complementary near-peers; the wide-gap forensic top-accuracy pair gains
only on a small-weight plateau ($x=1/12$--$1/6$). Finally, on \textbf{RQ3}, the heuristic calibrated once on SuperGPQA at
40:60 transfers as the most stable pre-pooling predictor.
For practitioners: measure the swap mass where pooled votes exist,
otherwise screen with $\hat S$ (Section~\ref{sec:samples}).

\bibliographystyle{ACM-Reference-Format}
\bibliography{refs}

\appendix
\raggedbottom
\setlength{\intextsep}{6pt plus 2pt minus 2pt}
\renewcommand{\topfraction}{0.9}\renewcommand{\textfraction}{0.05}\renewcommand{\floatpagefraction}{0.8}

\needspace{5\baselineskip}
\section{Data Availability}\label{app:data}

The \emph{deciban} corpus --- the per-vote inference records, vote counts,
grades, gold answers and abstention and error reasons for all three datasets
--- is available at:
\url{https://huggingface.co/datasets/IcyApril/deciban}

The forensic testbed --- the Apptainer sandbox, agent harness and offline
grader --- is open-source at:
\url{https://github.com/alan-turing-institute/deciban}

The forensic benchmark comprises five publicly available challenges: the
NIST CFReDS Data Leakage Case~\cite{nist2015dataleakage}, with answers from
NIST's published answer document; Rhino Hunt, the DFRWS 2005 Forensics Rodeo
challenge contributed by Golden G. Richard III and re-hosted by NIST
CFReDS~\cite{richard2005rhino}, with answers from the official DFRWS answer
document; the Malware-Traffic-Analysis.net exercises ``Lumma in the
Room-ah''~\cite{mta2026lumma} and ``Easy As 123''~\cite{mta2026easyas}, with
answers from the site's own answer pages; and the iOS image from the Magnet
Virtual Summit 2025 CTF, built by Hexordia and distributed via NIST
CFReDS~\cite{hexordia2025mvsctf}, with answers drawn from the community
writeups of Andro6~\cite{andro62025writeups} and
Metz~\cite{metz2025exploring}.

\section{Ethical Use of Data}\label{app:ethics}

The corpus involves no human participants and no personal data: all
inferences are model outputs over published benchmark questions and publicly
released digital-forensics challenges, used for research with attribution;
provenance and access records are cited in Appendix~\ref{app:data}. GPQA's authors ask that its examples not be
revealed publicly online~\cite{gpqa_dataset_card}; the corpus therefore
carries question identifiers and grades only.

\section{Additional Charts}\label{app:weights}

\begin{figure}[t]
  \centering
  \includegraphics[width=\columnwidth]{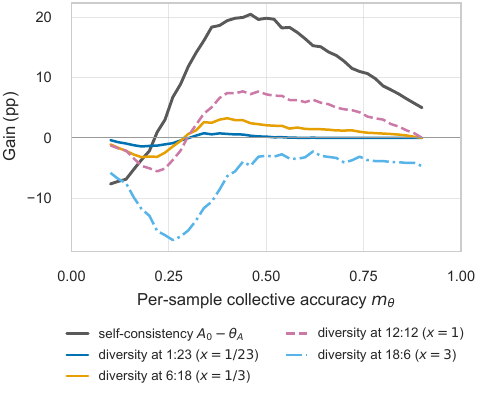}
  \caption{Decomposition of the pooled vote's gain over a single primary
  sample ($\Delta_\theta=0.10$, $\phiadj\approx 0.5$)}
  \Description{Self-consistency and diversity contributions to pooled-vote gain against collective accuracy.}
  \label{fig:decomp}
\end{figure}

\begin{figure}[t]
  \centering
  \includegraphics[width=\columnwidth]{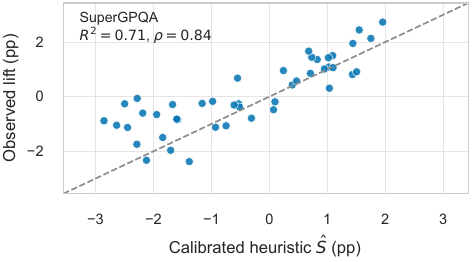}
  \caption{Calibrated heuristic against observed lift at 40:60, SuperGPQA}
  \Description{Scatter of calibrated heuristic against observed lift on SuperGPQA.}
  \label{fig:calibration}
\end{figure}

\begin{figure}[t]
  \centering
  \includegraphics[width=\columnwidth]{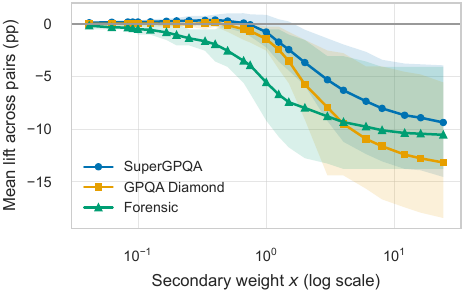}
  \caption{Mean lift across pairs against secondary weight (IQR shaded)}
  \Description{Mean lift across pairs against secondary weight for each dataset, with interquartile bands.}
  \label{fig:meanlift}
\end{figure}

\begin{figure}[t]
  \centering
  \includegraphics[width=\columnwidth]{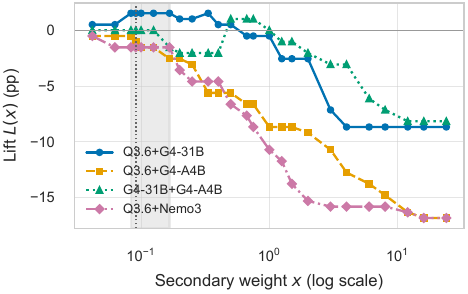}
  \caption{Lift against secondary weight for the strongest forensic pairs ($x=1/12$--$1/6$ plateau shaded)}
  \Description{Lift against secondary weight for the strongest forensic pairs with the small-weight plateau shaded.}
  \label{fig:cyber_weights}
\end{figure}

\begin{figure}[t]
  \centering
  \includegraphics[width=0.85\columnwidth]{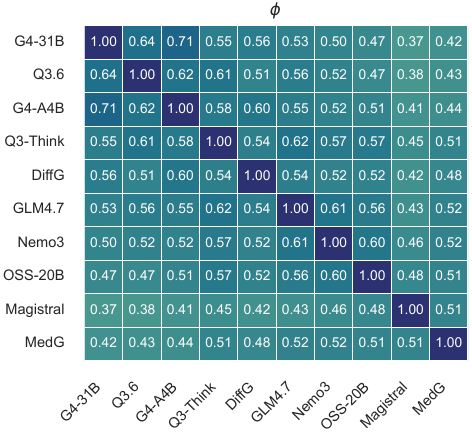}
  \caption{Correctness correlation $\phi$ across model pairs, SuperGPQA}
  \Description{Heatmap of raw correctness correlation for all SuperGPQA model pairs.}
  \label{fig:phi_sg}
\end{figure}

\begin{figure}[t]
  \centering
  \includegraphics[width=\columnwidth]{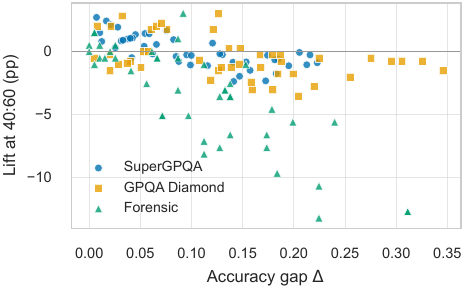}
  \caption{Observed lift at 40:60 against the accuracy gap, all pairs}
  \Description{Scatter of observed lift at the 40:60 split against accuracy gap for all pairs.}
  \label{fig:liftdelta}
\end{figure}

Figure~\ref{fig:decomp} decomposes the total gain the pooled vote offers
over a single primary sample into its two parts, in the question-clustered
world of Section~\ref{sec:simulation}. Repeated
sampling of the primary alone supplies the self-consistency gain
($A_0-\theta_A$): voting converts per-sample accuracy $0.95$ to
near-certainty on the questions the primary knows, forfeits the occasional
lucky sample on those it does not, and on contested questions below the
threshold amplifies distractors, so the gain runs from $-8$pp at the
foot to $+5$pp when the primary knows everything, crossing zero near
$m_\theta=0.22$. Diversity supplies the pair-dependent
increment on top ($A_x-A_0$), peaking mid-range where rescue opportunity
is greatest. The two parts thus dominate in different regimes ---
self-consistency where the primary is already strong, diversity where the
pair is balanced and complementary. Figure~\ref{fig:calibration} charts the
calibrated heuristic against observed lift on the calibration set.
Figure~\ref{fig:meanlift} plots mean lift across all 45 pairs against the
secondary weight, per dataset: the cross-pair mean peaks at $x=0.4$ on
SuperGPQA
and $x=2/3$ is the largest grid weight at which it remains positive
(Section~\ref{sec:calibration}), whilst the strongest forensic pairs'
curves (Figure~\ref{fig:cyber_weights}) peak on the small-weight plateau of
Section~\ref{sec:cyberlift} and collapse towards the $-\Delta$ limit past
equal weighting. Figure~\ref{fig:phi_sg} is the raw-$\phi$ counterpart of
Figure~\ref{fig:phiadj_sg}; comparing the two shows the systematic
depression of raw $\phi$ for wide-gap pairs that the accuracy adjustment of
Eq.~(\ref{eq:phiadjdef}) removes. Finally, Figure~\ref{fig:liftdelta}
charts observed lift at 40:60 against the accuracy gap across all pairs.

\balance
\section{Additional Tables}\label{app:pairs}

Table~\ref{tab:abstentions} reports per-model abstention rates: on the MCQ
corpora the dominant source is reasoning truncated at the token cap
(`Trunc.'), whilst on the forensic tasks it is questions left unanswered at
trace termination. Table~\ref{tab:trios} compares the best grid-searched
pair against the best grid-searched trio per dataset
(Section~\ref{sec:trios}), and Table~\ref{tab:top_pairs} lists the eight
SuperGPQA pairs with the largest observed lift at the 40:60 weighting,
alongside the calibrated heuristic $\hat S$ and the measured swap mass
$\Sstar$ for each.

\begin{table}[H]
\centering
\renewcommand{\arraystretch}{1.15}
\caption{Abstention (null-vote) rates by model (\%).}
\label{tab:abstentions}
\footnotesize
\begin{tabular}{|l|r|r|r|r|}
\hline
\textbf{Model} & \textbf{Parsed} & \textbf{Trunc.} & \textbf{No parse} & \textbf{Unans.} \\ \hline
Gemma-4-31B & 100.0 & 0.0 & 0.0 & 47.1 \\ \hline
Qwen3.6-27B & 97.7 & 2.2 & 0.1 & 25.3 \\ \hline
Gemma-4-26B-A4B & 99.2 & 0.7 & 0.1 & 68.2 \\ \hline
Qwen3-30B-Think. & 99.9 & 0.1 & 0.0 & 32.4 \\ \hline
DiffusionGemma-26B & 95.1 & 4.1 & 0.8 & 90.2 \\ \hline
GLM-4.7-Flash & 89.4 & 10.4 & 0.2 & 76.1 \\ \hline
Nemotron-3-Nano & 89.5 & 10.0 & 0.5 & 74.0 \\ \hline
GPT-OSS-20B & 98.6 & 0.2 & 1.2 & 96.6 \\ \hline
Magistral-Small & 98.9 & 0.2 & 1.0 & 47.2 \\ \hline
MedGemma-27B & 97.0 & 1.7 & 1.3 & 41.5 \\ \hline
\end{tabular}
\\[3pt]
{\footnotesize First three columns over 76,752 MCQ inferences per model; Trunc. = reasoning cut at the token cap. Unans. = share of the 2,352 gradeable forensic question instances with no submitted answer.}
\end{table}

\begin{table}[H]
\centering
\renewcommand{\arraystretch}{1.05}
\setlength{\tabcolsep}{3pt}
\caption{Best grid-searched pair against best grid-searched trio, per dataset.}
\label{tab:trios}
\footnotesize
\begin{tabular}{|l|l|l|r|r|}
\hline
\textbf{Dataset} & \textbf{Ensemble} & \textbf{$x$} & \textbf{Lift} & \textbf{Ceiling} \\ \hline
SuperGPQA & OSS-20B+MedG & 1 & +2.88 & 10.93 \\ \hline
SuperGPQA & OSS-20B+Magistral+MedG & 1/2, 2/3 & +3.68 & 17.24 \\ \hline
GPQA Diamond & OSS-20B+GLM4.7 & 2/5 & +4.38 & 7.83 \\ \hline
GPQA Diamond & OSS-20B+GLM4.7+Magistral & 1/2, 1/3 & +4.88 & 15.74 \\ \hline
Forensic & G4-31B+Q3-Think & 2/3 & +3.06 & 5.10 \\ \hline
Forensic & G4-A4B+Nemo3+Q3-Think & 1/4, 1/2 & +3.57 & 3.57 \\ \hline
\end{tabular}
\\[3pt]
{\footnotesize Lift and selection ceiling (rescue mass) in pp over the trio's primary; weights in 24ths of the primary's vote. In-sample maxima, not held-out estimates.}
\end{table}

\setlength{\tabcolsep}{2.5pt}
\begin{table}[H]
\centering
\renewcommand{\arraystretch}{1.15}
\caption{The eight SuperGPQA pairs with the largest observed lift at the 40:60 weighting.}
\label{tab:top_pairs}
\footnotesize
\begin{tabular}{|l|l|r|r|r|r|r|r|}
\hline
\textbf{Primary} & \textbf{Secondary} & \textbf{$p$ \%} & \textbf{$q$ \%} & \textbf{$\phi_{\mathrm{adj}}$} & \textbf{$\hat S$} & \textbf{$S^{\star}$} & \textbf{Lift} \\ \hline
Magistral & MedG & 47.4 & 46.7 & 0.51 & 1.96 & 2.28 & +2.73 \\ \hline
DiffG & GLM4.7 & 54.9 & 53.2 & 0.56 & 1.55 & 2.20 & +2.44 \\ \hline
OSS-20B & Magistral & 49.5 & 47.4 & 0.50 & 1.75 & 1.88 & +2.13 \\ \hline
OSS-20B & MedG & 49.5 & 46.7 & 0.54 & 1.44 & 1.66 & +1.95 \\ \hline
DiffG & Magistral & 54.9 & 47.4 & 0.48 & 0.68 & 1.49 & +1.67 \\ \hline
G4-31B & Q3.6 & 69.0 & 67.9 & 0.66 & 1.09 & 1.35 & +1.51 \\ \hline
DiffG & OSS-20B & 54.9 & 49.5 & 0.58 & 0.73 & 1.25 & +1.44 \\ \hline
GLM4.7 & Magistral & 53.2 & 47.4 & 0.49 & 1.02 & 1.02 & +1.43 \\ \hline
\end{tabular}
\\[3pt]
{\footnotesize $\hat S$, $S^{\star}$ and lift in percentage points. Model codes as in Figure~\ref{fig:phiadj_sg}.}
\end{table}
\setlength{\tabcolsep}{6pt}

\end{document}